\def\year{2020}\relax
\documentclass[letterpaper]{article} 
\usepackage{aaai20}  
\usepackage{times}  
\usepackage{helvet} 
\usepackage{courier}  
\usepackage[hyphens]{url}  
\usepackage{graphicx} 
\usepackage{graphicx}  
\usepackage{amsmath}
\usepackage{amsfonts}
\usepackage{subcaption}
\usepackage{booktabs} 

\title{Transparent Classification with \\ Multilayer Logical Perceptrons and Random Binarization}
\author{Zhuo Wang\textsuperscript{\rm 1}, Wei Zhang\textsuperscript{\rm 2}\thanks{Corresponding authors.}, Ning Liu\textsuperscript{\rm 1}, Jianyong Wang\textsuperscript{\rm 1}\footnotemark[1]\\ 
\textsuperscript{\rm 1}Department of Computer Science and Technology, Tsinghua University\\
\textsuperscript{\rm 2}School of Computer Science and Technology, Shanghai Key Laboratory of Trustworthy Computing,\\ East China Normal University\\
wang-z18@mails.tsinghua.edu.cn, \{zhangwei.thu2011, victorliucs\}@gmail.com, jianyong@tsinghua.edu.cn 
}

\begin{document}

\maketitle

\begin{abstract}
Models with transparent inner structure and high classification performance are required to reduce potential risk and provide trust for users in domains like health care, finance, security, etc. However, existing models are hard to simultaneously satisfy the above two properties. In this paper, we propose a new hierarchical rule-based model for classification tasks, named Concept Rule Sets (CRS), which has both a strong expressive ability and a transparent inner structure. To address the challenge of efficiently learning the non-differentiable CRS model, we propose a novel neural network architecture, Multilayer Logical Perceptron (MLLP), which is a continuous version of CRS. Using MLLP and the Random Binarization (RB) method we proposed, we can search the discrete solution of CRS in continuous space using gradient descent and ensure the discrete CRS acts almost the same as the corresponding continuous MLLP. Experiments on 12 public data sets show that CRS outperforms the state-of-the-art approaches and the complexity of the learned CRS is close to the simple decision tree. Source code is available at https://github.com/12wang3/mllp.

\end{abstract}

\section{Introduction}
Relying on strong ability of data modeling, machine learning, especially deep learning, becomes the main paradigm for decision-making systems \cite{goodfellow2016deep,doshi2017towards}.
The decision-making systems have widespread usage in important areas such as medicine, finance, politics, as well as law, where people need the explanations why decisions are made to ensure their safety and protect their rights \cite{goodman2016european,lipton2016mythos}. 
As a result, the demand for the transparency of machine learning methods is increasing, which is crucial for earning the trust of users \cite{doshi2017towards} and reducing potential risks and bugs \cite{chu2018exact}.
However, most of the machine learning models can hardly ensure good predictive ability and transparency at the same time, and sacrificing transparency for good performance could result in serious consequences.

One important notion of transparency is that each part of the model, including input, parameter, and calculation, etc., admits an intuitive explanation \cite{lipton2016mythos}. For example, each node in Decision Tree corresponds to a rule description, e.g., ``\textit{if} \#citations $>$ 300 \textit{then} a good paper''. The main reason for deep neural networks not being transparent models is that the activation value of each neuron is not explainable. Unlike the Boolean state, i.e., True or False, used in the rule-based model, the continuous activation value of deep neural networks is hard to associate with one actual meaning. In addition, ensemble models like Random Forest are not transparent as well for the decision is made by hundreds of models which are hard to explain as a whole.

Linear model and rule-based model are two widely used transparent models, however, they both suffer from low model capability. Linear models cannot fit non-linear data well because of the limitation of their model structures. Rule-based models (e.g., Decision Tree, Rule Set, and Rule List) have strong model expressivity that can fit both linear and non-linear data well. However, a fundamental limitation with these rule-based models is that they find the rules by employing various heuristic methods \cite{Quinlan:1993:CPM:583200,breiman2017classification,cohen1995fast} which may not find the globally best solution or a solution with close performance. In addition, the gradient descent method cannot be directly applied to the rule-based model learning for the discrete parameters.

Studies in recent years provide some solutions to improve model transparency in different aspects. 
Surrogate models \cite{frosst2017distilling,ribeiro2016should,selvaraju2017grad} try to use a simple model to fit a complex model globally or locally.
Then understanding the complex model by interpreting the simple model.
Hidden layers investigation \cite{yosinski2015understanding,zhou2018interpreting,zhang2017interpretable} aims to visualize and analyze the statuses of hidden neurons or features learned by hidden neurons.
All these methods improve the model transparency in specific scenarios. 
However, there is always a gap between the surrogate model and the complex model it aims to fit, and this inconsistency may have an influence on subsequent model analysis and understanding. The interpretation provided by hidden layer investigation is very intuitive, but it could hardly be quantitatively measured and applied, which limits the scope of application of this method.
One fundamental problem of these methods is that they all learn complex models first to obtain high classification performance, then try to interpret these learned complex models which are hard to understand and the interpretations may be inconsistent. Unlike these methods aiming to improve the transparency of complex models with high performance, we can try the opposite way to solve the problem, i.e., improving the performance of transparent models.

As mentioned above, rule-based models have both transparent inner structures and strong model expressivity but suffer from lacking an efficient optimization method because of the discrete parameters. If we can adopt gradient descent to learn rule-based models, we may obtain a model with both transparency and high performance.

In this paper, we propose a new hierarchical rule-based model, \textbf{Concept Rule Sets (CRS)} (see Figure~\ref{fig:CRS}) and its continuous version, \textbf{Multilayer Logical Perceptron (MLLP)}, which is a neural network with logical activation functions and constrained weights. We can use gradient descent to learn the discrete CRS via continuous MLLP. To ensure the discrete CRS and the continuous MLLP have almost the same behavior, we propose a new training method called \textbf{Random Binarization (RB)}, which binarizes a randomly selected subset of weights in MLLP during the training process to enable the MLLP to keep consistency with the corresponding CRS.

The main contributions of this paper are as follows:

\noindent (\romannumeral1) We propose a new hierarchical rule-based model, Concept Rule Sets, with strong model expressivity and ability to learn transparent data representation for classification tasks. The complete definition of CRS is also given.

\noindent (\romannumeral2) We propose a new neural network, Multilayer Logical Perceptron, and a new training method, Random Binarization to learn CRS efficiently using gradient descent. We also provide solutions to overcome the vanishing gradient problems that occur during training.

\noindent (\romannumeral3) Based on the transparent structure of CRS, we propose two simplification methods to reduce the model complexity.

\noindent (\romannumeral4) We conduct experiments on 12 public data sets to compare the classification performance and complexity of our model with other representative classification models.

\section{Related Work}
There are four classes of methods that are directly related to this work, i.e., rule-based model, surrogate model, hidden layers investigation and binary neural network.

\textbf{Rule-based Models} are considered to be interpretable because of their transparent inner structure. Decision tree, rule list, and rule set are the widely used structure in rule-based models.
\cite{letham2015interpretable,wang2017bayesian,yang2017scalable} leverage Bayesian frameworks to restrict and adjust model structure more reasonably. \cite{lakkaraju2016interpretable} learns interpretable decision sets by using independent if-then rules and a non-monotone submodular objective function. \cite{angelino2017learning} learns certifiably optimal rule lists and leverages algorithmic bounds and efficient data structures to speed up.

However, most existing rule-based models need frequent itemsets mining and/or long-time searching, which limits their applications. Moreover, it is hard for these interpretable rule-based models to get comparable performance with complex models like Random Forest.

\textbf{Surrogate Models} use simple models to fit or approximate complex models (e.g., deep neural networks) globally or locally, and explain the complex model by interpreting the simple model.

\cite{hinton2015distilling} proposed a distillation method that trains a relatively small neural network to predict the output of a large network, regarding as learning knowledge of the large network. To get a more transparent model by distillation methods, \cite{frosst2017distilling} took place the small neural network with a decision tree.

\cite{ribeiro2016should} proposed LIME to interpret any classifier by using a transparent model 
to fit the classifier locally. \cite{chu2018exact} developed OpenBox to transform a piecewise linear neural network into a set of linear classifiers which help interpret the network.

\cite{zhou2016learning} presented a method that maps the predicted class score back to the previous convolutional layer to generate the class activation maps (CAMs). \cite{selvaraju2017grad} proposed a Grad-CAM method that generalizes CAM by using the First-order Taylor-series approximation to approximate the part after the last convolutional layer.

The surrogate approaches can use complex models to get higher accuracy and use simple approximate models to get interpretation. However, inconsistencies always exist between an actual model and its surrogate model \cite{kim2017interpretable}, and there is no guarantee for the authenticity of the interpretation from the surrogate model because of these inconsistencies. 

\textbf{Hidden Layers Investigations} visualize and analyze statuses of hidden neurons or their learned features.

\cite{yosinski2015understanding} introduced two tools to visualize the activations and features at each layer of a trained convnet.
\cite{mahendran2015understanding} contributed a general framework to reconstruct the image by inverting representations to analyze the visual information contained in representations. \cite{zhou2018interpreting} proposed a Network Dissection method that interprets networks by providing meaningful labels to their individual units. 
\cite{zhang2017interpretable} proposed an interpretable CNN that each filter in a high convolutional layer is assigned with an object part automatically during the learning process. Based on the interpretable CNN, \cite{zhang2018interpreting} further proposed to use a decision tree to mine the decision mode memorized in fully-connected layers.

The interpretation provided by hidden layer investigation is intuitive with visualization, but it could hardly be quantitatively measured and applied. The main difference between hidden layers investigation and our method is hidden layers investigation tries to interpret learned data representation while our method aims to learn a specific and interpretable data representation.

\textbf{Binary Neural Networks} train a DNN with binary weights during the forward and backward propagations \cite{cour2015bconnect,cour2016binarized}. Although their weights are binary, they are still much more complex than the rule-based models. And it is harder to understand the operations of binary neural networks than logical operations. Moreover, the model capability of the binary neural network is also restricted.

\section{Concept Rule Sets}

\subsection{Notation Description}
Let $\mathcal{D}=\{(X_1,\mathbf{y}_1),\dots,(X_N,\mathbf{y}_N)\}$ denote the training data set with $N$ instances, where $X_i$ is the observed feature vector of the $i$-th instance with the $j$-th entry as $X_{i,j}$, and $\mathbf{y}_i$ is the corresponding class label, $i\in\{1, \dots, N\}$.
All feature values can be discrete or continuous, and all the classes are discrete values.
After the discretization and binarization of all the data features and class labels, $\mathcal{D}$ is converted into $\mathcal{D}'=\{(X'_1,Y'_1),\dots,(X'_N,Y'_N)\}$, where $X'_i \in \{0, 1\}^{J}$ is a binary feature vector with the size of $J$ and $Y'_i \in \{0, 1\}^{C}$ is a one-hot class label vector whose size is equal to $C$, the number of class labels. Let $A'$ denote the set of binary features, and $a'_j \in A'$ is the $j$-th binary feature. Throughout this paper, we use 1 (True) and 0 (False) to represent the two states of a Boolean variable. Therefore, each dimension of the binary feature vector and one-hot class label vector can be considered as a Boolean variable.

\begin{figure}
    \centering
    \includegraphics[width=0.45\textwidth]{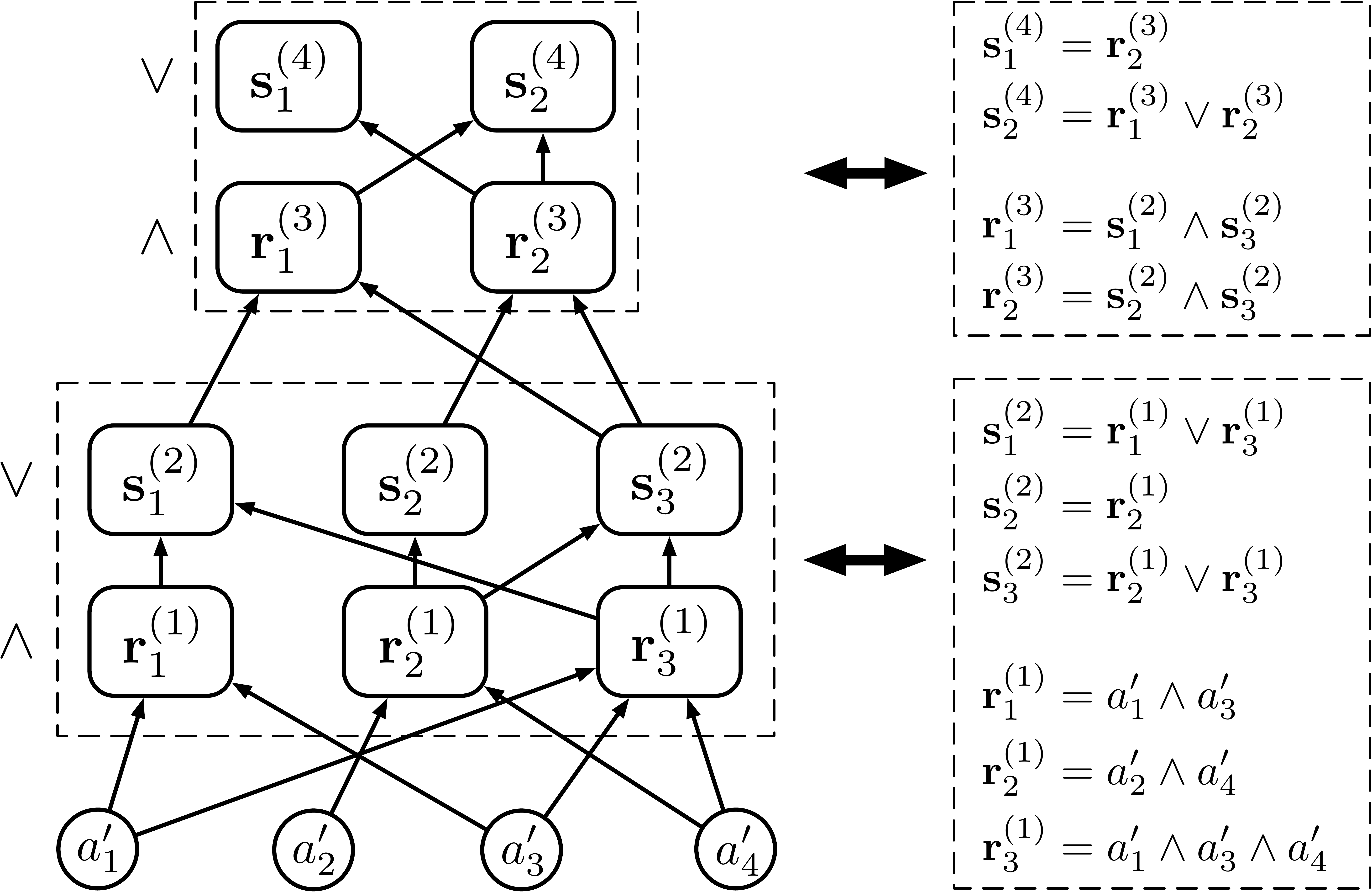}
    \caption{A Concept Rule Sets example for intuitive understanding. The left side and the right side are two forms of CRS. One dashed box corresponds to one level in CRS. Arrows of the edges indicate the flow of data.}
    \label{fig:CRS}
\end{figure}

\subsection{Definition and Application of CRS}
Before giving the definition of Concept Rule Sets, we first formulate the concepts of rule and rule set respectively.
A rule $\mathbf{r}_i$ is a conjunction of one or more Boolean variables. 
$$\mathbf{r}_i=\mathbf{b}_{i_1} \wedge \mathbf{b}_{i_2} \wedge \dots \wedge \mathbf{b}_{i_p}$$
where $\mathbf{b}_{i_k}$ ($k\in\{1,\dots,p\}$) is the Boolean variable in rule $\mathbf{r}_i$ and $p$ is the length of the rule $\mathbf{r}_i$. 
A rule set $\mathbf{s}_j$ is a disjunction of one or more rules, i.e., Disjunctive Normal Form (DNF).
$$\mathbf{s}_j=\mathbf{r}_{j_1} \vee \mathbf{r}_{j_2} \vee \dots \vee \mathbf{r}_{j_q}$$
where $\mathbf{r}_{j_k}$ ($k\in\{1,\dots,q\}$) is the rule in rule set $\mathbf{s}_j$ and $q$ is the number of rules in $\mathbf{s}_j$. 

A Concept Rule Sets (CRS), denoted by $\mathcal{F}$, is a multi-level structure in which each level contains a conjunction layer followed by a disjunction layer. 
For a CRS with $L$ levels, we denote its $l$-th layer ($l\in \{1,3,\dots, 2L-1\}$) by $\mathcal{R}^{(l)}$ for being a conjunction layer and denote the $l$-th layer ($l\in \{2,4,\dots, 2L\}$) by $\mathcal{S}^{(l)}$ for being a disjunction layer. 
For ease of expression, we represent the input layer, i.e., $0$-th layer, by $\mathcal{S}^{(0)}$.
Each layer in CRS contains a specific number of nodes and the edges connected with its previous layer, except $\mathcal{S}^{(0)}$. Let $\mathbf{n}_{l}$ denote the number of nodes in the $l$-th layer, and $W^{(l)}$ denote an $\mathbf{n}_{l}$-by-$\mathbf{n}_{l-1}$ adjacency matrix of the $l$-th layer and the $(l-1)$-th layer, where $W_{i,j}^{(l)} \in \{0,1\}$. $W_{i,j}^{(l)}=1$ indicates there exists an edge connecting the $i$-th node in $l$-th layer to the $j$-th node in $(l-1)$-th layer, otherwise $W_{i,j}^{(l)}=0$. Similar to neural networks, we regard these adjacency matrices as the weight matrices of CRS.

We denote the $i$-th node in layer $\mathcal{R}^{(l)}$ ($l\in \{1,3,\dots, 2L-1\}$) by $\mathbf{r}_{i}^{(l)}$, and the $i$-th node in layer $\mathcal{S}^{(l)}$ ($l\in \{0,2,\dots, 2L\}$) by $\mathbf{s}_{i}^{(l)}$.
Specifically speaking, node $\mathbf{r}_{i}^{(l)}$ corresponds to one rule, in which the Boolean variables are nodes in the previous layer connected with $\mathbf{r}_{i}^{(l)}$, 
while node $\mathbf{s}_{i}^{(l)}$ corresponds to one rule set, in which the rules are nodes in previous layer connected with $\mathbf{s}_{i}^{(l)}$.
Formally, the two types of nodes are defined as follows:
\begin{equation}
\mathbf{r}_{i}^{(l)} =
\bigwedge_{W_{i,j}^{(l)}=1}\mathbf{s}_{j}^{(l-1)},\;\;\;\;\;\; \mathbf{s}_{i}^{(l+1)} = \bigvee_{W_{i,j}^{(l+1)}=1}\mathbf{r}_{j}^{(l)}.
\end{equation}
A concrete example of CRS is shown in Figure \ref{fig:CRS}.

For any given instance, after setting the values of $\mathcal{S}^{(0)}$ according to its binary feature vector, we can compute the values of all the nodes in CRS layer by layer.
If we set the number of nodes in the last layer, i.e., $\mathbf{n}_{2L}$, to $C$, CRS can work as a classifier $\mathcal{F}:\{0,1\}^{J} \rightarrow \{0,1\}^{C}$, which outputs the values of nodes in the last layer $\mathcal{S}^{(2L)}$, and $\mathbf{s}_{i}^{(2L)}=1$ indicates that the CRS classifies the input instance as the $i$-th class label. If more than one dimension of $\mathbf{s}^{(2L)}$ has value equals to 1, we choose the first one as the result. Meanwhile, a crucial byproduct of CRS is the learned layer-wise representation of each instance, similar to the mechanism of deep neural networks \cite{goodfellow2016deep} but more transparent.
Let $\mathbf{h}^{(l)}$ denote the data representation learnt by the $l$-th layer in CRS, which is a binary vector with $\mathbf{n}_l$ dimensions.
\begin{equation*}
\mathbf{h}^{(l)}=
\begin{cases}
[\mathbf{r}_{1}^{(l)},\mathbf{r}_{2}^{(l)},\dots, \mathbf{r}_{\mathbf{n}_{l}}^{(l)}]^{\top}&  l\in \{1,3,\dots, 2L-1\} \\
[\mathbf{s}_{1}^{(l)},\mathbf{s}_{2}^{(l)},\dots, \mathbf{s}_{\mathbf{n}_{l}}^{(l)}]^{\top}& l\in \{2,4,\dots, 2L\}
\end{cases}
\end{equation*}
The value of $\mathbf{h}_{i}^{(l)}$ is equal to the value of the $i$-th node in the $l$-th layer which corresponds to a rule or a rule set. We can understand each dimension of $\mathbf{h}^{(l)}$ by analyzing the corresponding rule or rule set.
It is much easier than analyzing the real-valued weights and activation values of hidden neurons in deep neural networks.

The CRS model enjoys two major intrinsic advantages. One is that the model has strong expressivity because one level in CRS is equal to the widely used rule sets model which can fit both the linear and nonlinear data appropriately. The multi-level structure enhances the expressivity of CRS further. The other advantage is that the CRS model has a transparent inner structure and is able to learn transparent data representations, as aforementioned. 

However, how to search the appropriate weight matrices in CRS to obtain high classification performance and low model complexity remains a challenging problem.
We will discuss our proposed solution in the next section.

\section{Multilayer Logical Perceptron}
In this section, we introduce how to use Multilayer Logical Perceptron and Random Binarization method to learn CRS and then simplify the learned CRS.
\subsection{Data Discretization and Binarization}\label{sec:dis-bin}
For the reason that CRS only takes binary vector input and rules containing continuous features are hard to understand, we apply the recursive minimal entropy partitioning algorithm \cite{dougherty1995supervised} to discretize feature values.
This algorithm partitions one feature recursively by searching the partition boundary which minimizes the class information entropy of candidate partitions. Minimal Description Length Principle is used to determine the stopping criteria.
After data discretization, we use one-hot encoding to covert all the discrete features into binary features.
\begin{figure}
  \begin{subfigure}[c]{0.26\columnwidth}
    \includegraphics[width=\linewidth]{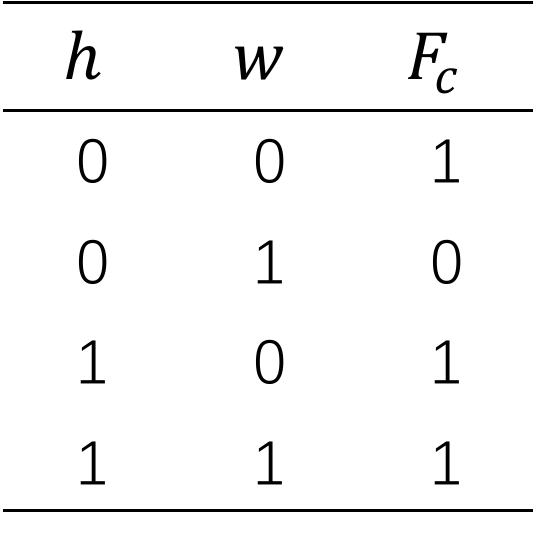}
    \caption{}
    \label{fig:Fc}
  \end{subfigure}
  \hfill 
  \begin{subfigure}[c]{0.26\columnwidth}
    \includegraphics[width=\linewidth]{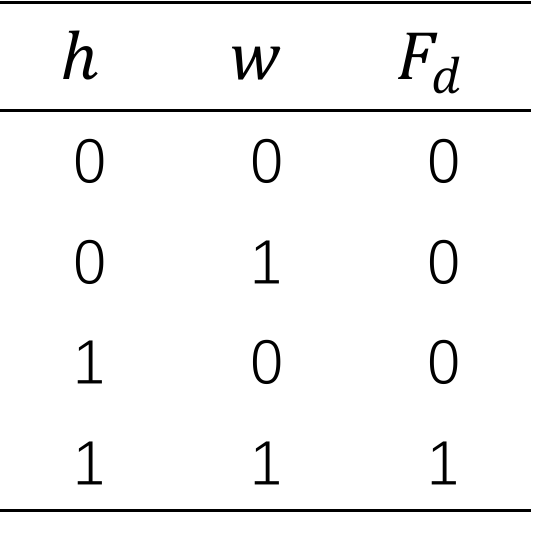}
    \caption{}
    \label{fig:Fd}
  \end{subfigure}
  \begin{subfigure}[c]{0.44\columnwidth}
    \includegraphics[width=\linewidth]{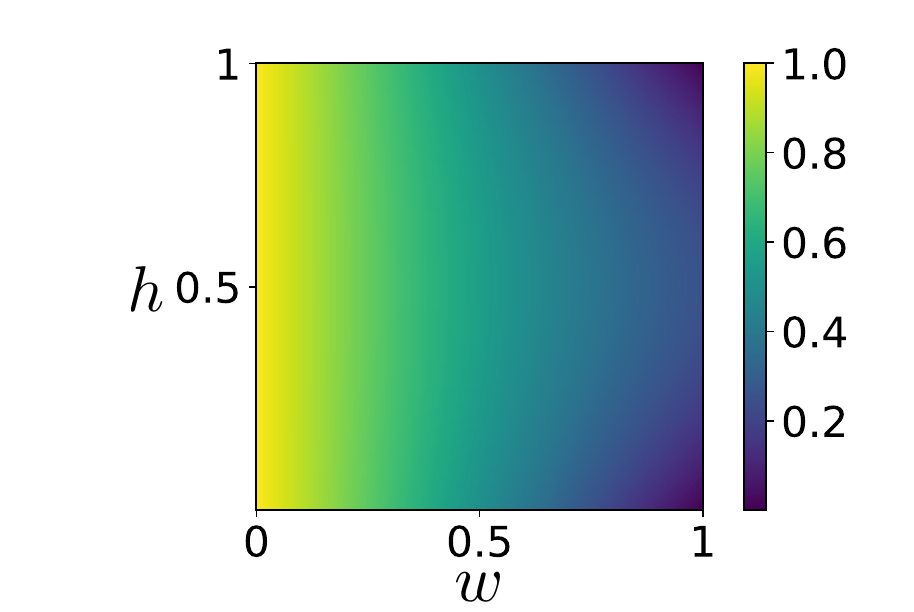}
    \caption{}
    \label{fig:Fd_Fc_heatmap}
  \end{subfigure}
\caption{(a) Truth table of $F_c(\cdot)$. (b) Truth table of $F_d(\cdot)$. (c) Heatmap of $F_c(h,w)\cdot (1-F_d(h,w))$.}
\label{fig:selector}
\end{figure}

\subsection{Network Structure and Training Method}
For the similarity between the structure of CRS and multilayer perceptron (MLP), a straightforward idea is to use the gradient descent method to train CRS, like the way of training MLP. However, we cannot compute the gradient of CRS due to the discrete weights. To overcome this problem, we propose a new neural network called Multilayer Logical Perceptron (MLLP) and a tailored training method, which can search the discrete solution of CRS in a continuous space using gradient descent.

MLLP and its corresponding training method aim at tackling the following three challenges:
\begin{itemize}
    \item Logical Activation Function. Commonly used activation functions cannot simulate the conjunction and disjunction operations.
    \item Vanishing Gradient Problem. The gradient could be extremely small due to the conjunction and disjunction operations.
    \item Discrete CRS Extraction. There is no guarantee the solutions found in continuous space could still work after converting them into discrete values.
\end{itemize}
\subsubsection{Overall Structure}
MLLP has $2L+1$ layers, same as the CRS model. And one neuron in MLLP corresponds to one node in CRS. Therefore, the number of neurons in $l$-th layer of MLLP is also $\mathbf{n}_{l}$. We denote the neuron corresponding to CRS node $\mathbf{r}_{i}^{(l)}$ by $\hat{\mathbf{r}}_{i}^{(l)}$ ($l\in\{1,3,\dots,2L-1\}$), and $\mathbf{s}_{i}^{(l)}$ by $\hat{\mathbf{s}}_{i}^{(l)}$ ($l\in\{2,4,\dots,2L\}$). All the layers in MLLP are fully connected layers. We denote the weights of the $l$-th layer of MLLP by $\hat{W}^{(l)}$, and $\hat{W}_{i,j}^{(l)}\in [0,1]$.

\subsubsection{Logical Activation Function}
In order to ensure the neurons in MLLP have the same behaviors as the corresponding nodes in CRS, activation functions that can simulate conjunction and disjunction operations are required. We adopt the conjunction function and disjunction function proposed in~\cite{payani2019learning}.
Specifically, given the two $n$-dimensional vectors $\mathbf{h}$ and $W_{i}$, the conjunction function \textit{Conj}$(\cdot)$ and disjunction function \textit{Disj}$(\cdot)$ are given by:
\begin{equation}
\label{eq:conj}
\begin{split}
\textit{Conj}(\mathbf{h}, W_{i})=\prod_{j=1}^{n}F_{c}(\mathbf{h}_{j}, W_{i,j}),
\end{split}
\end{equation} 
\begin{equation}
\label{eq:disj}
\begin{split}
\textit{Disj}(\mathbf{h}, W_{i})=1-\prod_{j=1}^{n}(1-F_{d}(\mathbf{h}_{j}, W_{i,j})),
\end{split}
\end{equation}
where $F_{c}(h,w)=1-w(1-h)$ and $F_{d}(h,w)=h\cdot w$.

In Equations \ref{eq:conj} and \ref{eq:disj}, if  $\mathbf{h}$ and $W_{i}$ are both binary vectors, then $\textit{Conj}(\mathbf{h}, W_{i})=\bigwedge_{W_{i,j}=1}\mathbf{h}_{j}$ and $\textit{Disj}(\mathbf{h}, W_{i})=\bigvee_{W_{i,j}=1}\mathbf{h}_{j}$. The truth tables, shown in Figure \ref{fig:selector}, indicate that $F_c(h,w)$ and $F_d(h,w)$ work as selectors, and they only select $\mathbf{h}_j$ to participate the operation when $W_{i,j}=1$. In other words, they replace $\mathbf{h}_j$ with the value that cannot affect the final results when $W_{i,j}=0$.

We apply the conjunction and disjunction functions to the neurons in MLLP as follows:
\begin{equation}
\label{eq:conj_hat}
\hat{\mathbf{r}}_{i}^{(l)} =\textit{Conj}(\hat{\mathbf{s}}^{(l-1)}, \hat{W}_{i}^{(l)}),\; l\in \{1,3,\dots, 2L-1\},
\end{equation} 
\begin{equation}
\label{eq:disj_hat}
\hat{\mathbf{s}}_{i}^{(l)}=\textit{Disj}(\hat{\mathbf{r}}^{(l-1)}, \hat{W}_{i}^{(l)}),\; l\in \{2,4,\dots, 2L\}.
\end{equation}
In order to maintain the characteristics of conjunction and disjunction functions, we need to constrain all the weights of MLLP in the range $[0, 1]$. A common approach is to replace $\hat{W}_{i,j}^{(l)}$ with $sigmoid(\hat{W}_{i,j}^{(l)})$ or $\frac{1}{2}(tanh(\hat{W}_{i,j}^{(l)})+1)$. However, these constraint functions keep the weights from being exact 0 or 1, and cause the vanishing gradient problem, for which we will discuss the reason later. To overcome this problem, we propose to use \textit{Clip} function to clip the weights after updating them with gradients, given as:
\begin{equation}
Clip(\hat{W}_{i,j}^{(l)})=Max(0, Min(1, \hat{W}_{i,j}^{(l)})).
\end{equation}

So far, MLLP has the ability to act exactly the same as the corresponding CRS when their weights are equal to the same discrete values, and MLLP is still differentiable.

\subsubsection{Loss Function}
The MLLP is denoted by $\hat{\mathcal{F}}$ and parameterized by $\hat{\mathcal{W}}$ including all the weights in $\hat{\mathcal{F}}$. The loss function for training is given by:
\begin{equation}
Loss=\frac{1}{N}\sum_{i=1}^{N}\textit{MSE}(Y'_{i},\hat{\mathcal{F}}(X'_i, \hat{\mathcal{W}}))+\lambda \Omega(\hat{\mathcal{W}}),
\end{equation}
where $\textit{MSE}(\cdot)$ is Mean Squared Error (MSE) and $\Omega(\hat{\mathcal{W}})$ is the L2 regularization.
The MSE criterion aims to minimize the gaps between the continuous output vector and one-hot label vector in each dimension, which could benefit the discrete CRS extraction. The L2 regularization encourages the MLLP to search for a CRS model with shorter rules.

\subsubsection{Vanishing Gradient Problem}
MLLP suffers from vanishing gradient problem during training and fails to converge. There are two main reasons for this problem: 

One reason is the constraint functions mentioned above. Let $\sigma(\cdot)$ denote these constraint functions. The derivative of $\sigma(\hat{W}_{i,j}^{(l)})$ is close to 0 when $\sigma(\hat{W}_{i,j}^{(l)})$ is close to 0 or 1.
However, the weights of CRS that MLLP aims to learn are 0 or 1, which means derivatives close to 0 are inevitable. 
The \textit{Clip} function we propose to use has no influence on the gradient and only clips the weights after weight updating. Therefore, the \textit{Clip} function would not cause the vanishing gradient problem. We have found that using the \textit{Clip} function instead of constraint functions is indeed beneficial for model convergence in practice.

The other reason for the vanishing gradient problem can be found by analyzing the partial derivative of each neuron w.r.t. its directly connected weights as follows:
\begin{equation}
\label{eq:derivative1}
\frac{\partial \hat{\mathbf{r}}_{i}^{(l)}}{\partial \hat{W}_{i,j}^{(l)}}
=(\hat{\mathbf{s}}_{j}^{(l-1)}-1) \cdot \prod_{k \neq j}F_{c}(\hat{\mathbf{s}}_{k}^{(l-1)}, \hat{W}_{i,k}^{(l)}),
\end{equation}
\begin{equation}
\label{eq:derivative2}
\frac{\partial \hat{\mathbf{s}}_{i}^{(l)}}{\partial \hat{W}_{i,j}^{(l)}}
=\hat{\mathbf{r}}_{j}^{(l-1)} \cdot \prod_{k \neq j}(1-F_{d}(\hat{\mathbf{r}}_{k}^{(l-1)}, \hat{W}_{i,k}^{(l)})).
\end{equation}
Due to the values of inputs and weights are in the range $[0, 1]$, the values of $F_c(\cdot)$ and $F_d(\cdot)$ in Equations \ref{eq:derivative1} and \ref{eq:derivative2} are in the range $[0, 1]$ as well. If $\mathbf{n}_l$ is large and most of the values of $F_c(\cdot)$ or $(1-F_d(\cdot))$ are not close to 1, then the derivative is close to 0 because of the multiplications.
As such, if we randomly initialize weights in range $[0, 1]$, the derivatives could be extremely small, especially when the size of model is very large. The heatmap of values about $F_c(h,w)\cdot (1-F_d(h,w))$ is shown in Figure \ref{fig:Fd_Fc_heatmap}.
We can observe that if $w$ is close to 0, then both $F_c(h,w)$ and $(1-F_d(h,w))$ are colse to 1. Therefore, a simple but efficient solution is to initialize the weights with small values close to 0.
Actually, we randomly initialize the weights with the distribution $\textit{Uniform}(0, 0.1)$.
\subsubsection{Discrete CRS Extraction}
After traning the MLLP, we need to extract the discrete CRS from it. A straightforward approach is to binarize all the weights using a threshold. The function for binarization could be:
\begin{equation}
Binarize(w, \mathcal{T})=\mathbb{I}(w > \mathcal{T}),
\end{equation} 
where $\mathbb{I}(\cdot)$ is the 0-1 indicator function and $\mathcal{T} \in (0, 1)$ is the threshold for binarization. We usually set $\mathcal{T}$ to 0.5.

However, the behavior of extracted CRS is very different from MLLP, and the extracted CRS can hardly be used for classification, especially when MLLP is deep. This situation is also shown in the experimental results of Table \ref{tab:accuracy}. The reason for this problem is that the neurons do not play the role of conjunction and disjunction operators as we expected due to the real-valued weights. 

To tackle this problem, we propose a new training method called Random Binarization (RB). When using the RB method during training, we randomly select a subset of weights in MLLP and binarize these weights. We fix these binary weights and only update other weights at each step. After several steps, we set these selected weights to their original values before binarization.
Then we select a new subset of weights and repeat the above procedure.
Let $M^{(l)}$ denote the mask matrix of $\hat{W}^{(l)}$, where $M_{i,j}^{(l)} \in \{0,1\}$ and $M_{i,j}^{(l)}=\mathbb{I}(p<\mathcal{P})$
with $p \sim \textit{Uniform}(0, 1)$ and $\mathcal{P}$ as the rate of binarization. Let $\tilde{W}_{i,j}^{(l)}$ denote the weight after random binarization, which is given as:
\begin{equation}
\tilde{W}_{i,j}^{(l)} =
\begin{cases}
\text{$\hat{W}_{i,j}^{(l)}$}& \text{$M_{i,j}^{(l)}=0$},\\
\text{$Binarize(\hat{W}_{i,j}^{(l)}, \mathcal{T})$}& \text{$M_{i,j}^{(l)}=1$}.
\end{cases}
\end{equation}
$\hat{W}_{i}^{(l)}$ in Equation \ref{eq:conj_hat} and \ref{eq:disj_hat} is now replaced by $\tilde{W}_{i}^{(l)}$. To fix the values of selected weights, we compute the partial derivative of $\tilde{W}_{i,j}^{(l)}$ w.r.t. $\hat{W}_{i,j}^{(l)}$ by $\frac{\partial \tilde{W}_{i,j}^{(l)}}{\partial \hat{W}_{i,j}^{(l)}}=1-M_{i,j}^{(l)}$,
and update the weights based on the following formula,
\begin{equation}
\frac{\partial Loss}{\partial \hat{W}_{i,j}^{(l)}}=\frac{\partial Loss}{\partial \tilde{W}_{i,j}^{(l)}} \cdot \frac{\partial \tilde{W}_{i,j}^{(l)}}{\partial \hat{W}_{i,j}^{(l)}}=\frac{\partial Loss}{\partial \tilde{W}_{i,j}^{(l)}} \cdot (1-M_{i,j}^{(l)}).
\end{equation}

We use $W_{i,j}^{(l)}=Binarize(\hat{W}_{i,j}^{(l)}, \mathcal{T})$  to extract a discrete CRS from the MLLP trained by the RB method. 
Thus the behaviors of CRS are almost the same as MLLP, which has a good classification performance.

Moreover, the RB method also works as the Dropout regularization \cite{srivastava2014dropout} which significantly reduces overfitting. The reason is that those weights binarized to 0 could be considered to be removed from the model, which is similar to randomly dropping neurons (along with their connections) as the Dropout regularization does.

\subsection{CRS Simplification}
Benefiting from the transparent inner structure of CRS, we simplify CRS by analyzing each node. 
Specifically, we propose two methods to simplify the CRS model extracted from MLLP:
\subsubsection{Dead Nodes Detection}
We name a node $v$ in CRS as ``dead node" if there exists no path from the top layer to the bottom layer that contains node $v$ or inputting all the training data into CRS cannot activate node $v$.
We can delete these dead nodes without affecting the performance of CRS.

A node $v$ in CRS is activated when the value of $v$ is 1 and inactivated when the value is 0. 
We are able to know whether a node is activated clearly, which is very difficult to distinguish the boundaries in common neural networks because of real-valued activation values. 

\begin{table*}[!h]
  \caption{5-fold cross validated F1 score (Macro) of each comparing algorithm on 12 UCI data sets.}
  \label{tab:accuracy}
\scalebox{0.78}{
  \begin{tabular}{c|cccccccccccc}
    \toprule
    Dataset & \textbf{CRS} & MLLP($\mathcal{P}$=0) & CRS($\mathcal{P}$=0) & C4.5 & CART & SBRL & LR & SVM & PLNN(MLP) & GBDT & RF(e=10) & RF(e=100)\\ 
    \midrule
adult & \textbf{80.95} & 74.59 & 51.39 & 75.40 & 74.77 & 79.88 & 78.43 & 63.63 & 73.55 & 80.36 & 77.48 & 78.83\\ 
bank-marketing & 73.34 & 63.38 & 46.88 & 71.24 & 70.21 & 72.67 & 69.81 & 66.78 & 72.40 & \textbf{75.28} & 69.89 & 72.01\\ 
banknote & 94.93 & 93.29 & 88.68 & 98.45 & 97.85 & 94.44 & 98.82 & \textbf{100.00} & \textbf{100.00} & 99.48 & 99.11 & 99.19\\ 
blogger & \textbf{85.33} & \textbf{85.33} & 20.00 & 75.90 & 78.27 & 67.64 & 55.55 & 62.11 & 56.24 & 67.58 & 77.33 & 85.17\\ 
chess & 80.21 & \textbf{80.49} & 71.56 & 79.90 & 79.15 & 26.44 & 33.06 & 36.83 & 77.85 & 71.41 & 66.38 & 74.25\\ 
connect-4 & \textbf{65.88} & 56.35 & 26.71 & 61.66 & 61.24 & 48.54 & 49.87 & 50.17 & 64.55 & 64.45 & 61.95 & 62.72\\ 
letRecog & 84.96 & 81.26 & 40.32 & 88.20 & 87.62 & 64.32 & 72.05 & 74.90 & 92.34 & \textbf{96.51} & 93.61 & 96.15\\ 
magic04 & 80.87 & 82.25 & 39.24 & 80.31 & 80.05 & 82.52 & 75.72 & 75.64 & 83.07 & \textbf{86.67} & 84.90 & 86.48\\ 
mushroom & \textbf{100.00} & \textbf{100.00} & 34.40 & \textbf{100.00} & 99.98 & \textbf{100.00} & \textbf{100.00} & \textbf{100.00} & \textbf{100.00} & \textbf{100.00} & \textbf{100.00} & \textbf{100.00}\\ 
nursery & \textbf{99.69} & 98.02 & 12.69 & 95.55 & 95.47 & 71.32 & 64.64 & 82.48 & 79.71 & 87.32 & 88.43 & 90.31\\ 
tic-tac-toe & \textbf{99.77} & \textbf{99.77} & 38.06 & 91.70 & 94.21 & 98.39 & 98.12 & 98.07 & 98.26 & 99.19 & 94.85 & 98.37\\ 
wine & 97.78 & 98.10 & 97.78 & 95.48 & 94.39 & 95.84 & 95.16 & 96.05 & 76.07 & \textbf{98.44} & 96.90 & 98.31\\ 
\midrule
\textbf{Average} & \textbf{86.98} & 84.40 & 47.31 & 84.48 & 84.44 & 75.17 & 74.27 & 75.56 & 81.17 & 85.56 & 84.24 & 86.82\\
  \bottomrule
\end{tabular}
}
\end{table*}

\begin{table}[!h]
  \caption{Data sets properties.}
  \label{tab:property}
  \scalebox{0.85}{
  \begin{tabular}{cccp{0.15\columnwidth}p{0.15\columnwidth}}
    \toprule
    Dataset & \#instances & \#classes	& \#original features	& \#binary features\\
    \midrule
adult & 32561 & 2 & 14 & 155\\ 
bank-marketing & 45211 & 2 & 16 & 88\\ 
banknote & 1372 & 2 & 4 & 17\\ 
blogger & 100 & 2 & 5 & 15\\ 
chess & 28056 & 18 & 6 & 40\\ 
connect-4 & 67557 & 3 & 42 & 126\\ 
letRecog & 20000 & 26 & 16 & 155\\ 
magic04 & 19020 & 2 & 10 & 79\\ 
mushroom & 8124 & 2 & 22 & 117\\ 
nursery & 12960 & 5 & 8 & 27\\ 
tic-tac-toe & 958 & 2 & 9 & 27\\ 
wine & 178 & 3 & 13 & 37\\
  \bottomrule
\end{tabular}
}
\end{table}

\subsubsection{Redundant Rules Elimination}
The redundant rules in CRS are useless, resulting the model to be more complex. Two typical examples of redundant rules are shown in Figure \ref{fig:CRS}. $\mathbf{s}_{1}^{(2)}=\mathbf{r}_{1}^{(1)} \vee \mathbf{r}_{3}^{(1)}$ can be simplified as $\mathbf{s}_{1}^{(2)}=\mathbf{r}_{1}^{(1)}$ for the weights of $\mathbf{r}_{1}^{(1)}$ is the subset of the weights of $\mathbf{r}_{3}^{(1)}$. Similarly, $\mathbf{r}_{2}^{(3)}=\mathbf{s}_{2}^{(2)} \wedge \mathbf{s}_{3}^{(2)}$ can be simplified as $\mathbf{r}_{2}^{(3)}=\mathbf{s}_{2}^{(2)}$. We define the subset check function of the weights by
\begin{equation}
\textit{Subset}(W_{i}^{(l)}, W_{j}^{(l)})=\mathbb{I}(\forall W_{i,k}^{(l)}=1, W_{j,k}^{(l)}=1).
\end{equation}
For $l\in \{2,3,\dots,2L\}$, the set of redundant weights that can be simplified is:
\begin{equation}
\{W_{i,j}^{(l)}\;\big|\; \exists k, \textit{Subset}(W_{k}^{(l-1)},W_{j}^{(l-1)})\wedge W_{i,k}^{(l)}=1\}.
\end{equation}

\section{Experiments}
In this section, we conduct experiments to evaluate the proposed method and answer the following questions:

\begin{enumerate}
    \item  How is the classification performance of CRS compared to other state-of-the-art models?
    
    \item How does the binarization rate $\mathcal{P}$ affect CRS?
    
    \item How is the model complexity of CRS?
\end{enumerate}

\subsection{Dataset Description}
We took 12 datasets from the UCI machine learning repository \cite{Dua:2019}, all of which are often used to test classification performance and model transparency \cite{letham2015interpretable,wang2017bayesian,yang2017scalable,huhn2009furia}.
Table \ref{tab:property} summarizes the statistics of these 12 datasets.
Together they show the data diversity, ranging from 100 to 67557 instances, from 2 to 26 classes, and from 4 to 42 original features.

\begin{figure}
    \centering
    \includegraphics[width=0.4\textwidth]{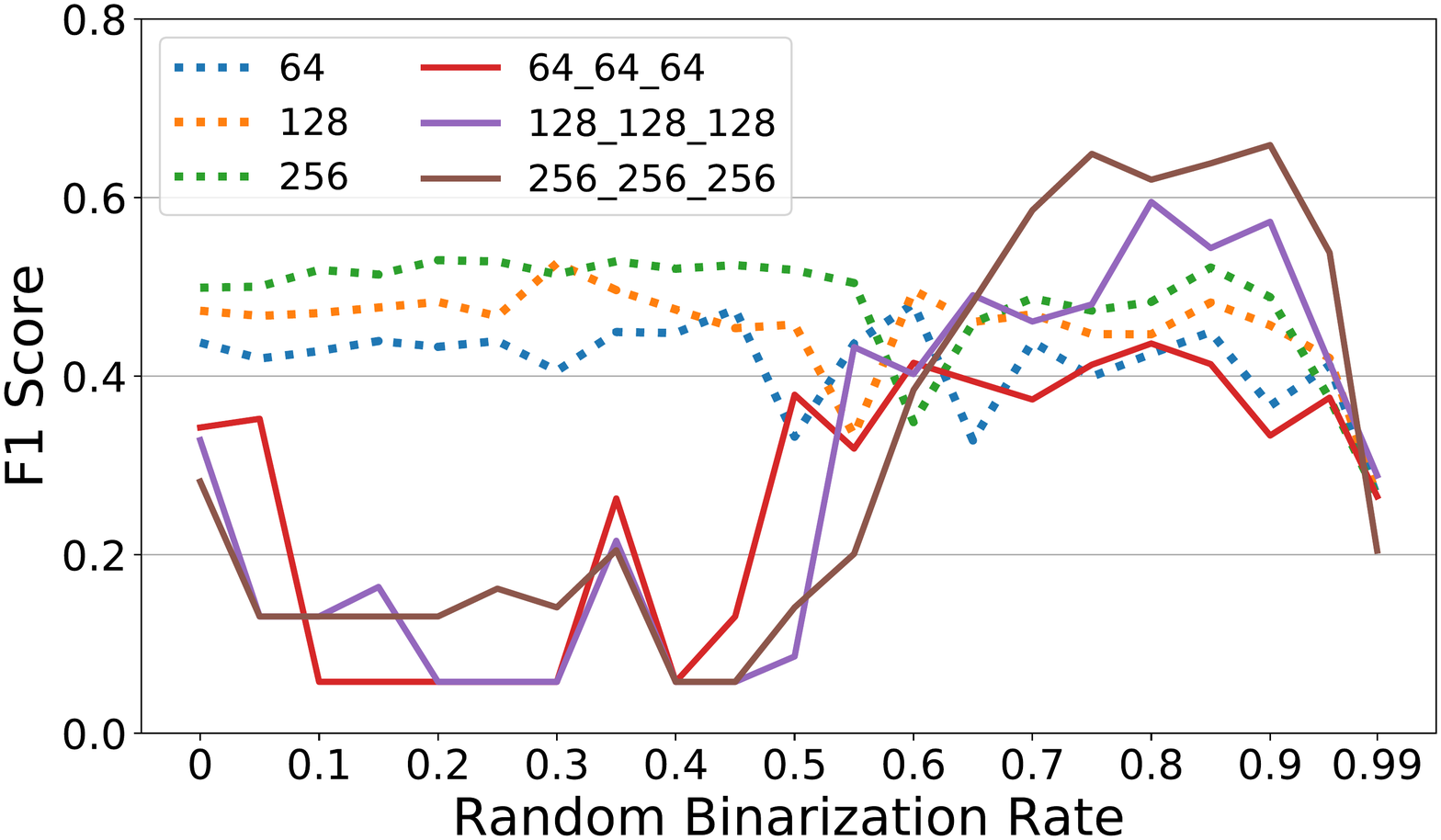}
    \caption{The impact of different binarization rates on CRS with different structures trained on Connect-4 data set.}
    \label{fig:rb_rate}
\end{figure}
\subsection{Experimental Settings}
\subsubsection{Evaluation Protocols}
Considering that some of the data sets are imbalanced, i.e., the number of different classes are quite different, we adopt the F1 score (Macro) as the classification performance metric. To evaluate the classification performance of our model and baselines more fairly, 5-fold cross-validation is adopted to have a lower bias on experimental results. Additionally, 80\% of the training set is used for training and 20\% for validation when parameters tuning is required. The total length of all rules is a commonly used metric for model complexity of rule-based models. However, in some of the rule-based models, there are lots of reused structures, e.g., one branch in Decision Tree can correspond to several rules, and the total length of all rules may be not fair for these models. Considering the fact that edges in Decision Tree and CRS determine the final model structure, we use the total number of edges in the model to measure model complexity.
\begin{figure*}
    \centering
    \includegraphics[width=0.95\textwidth]{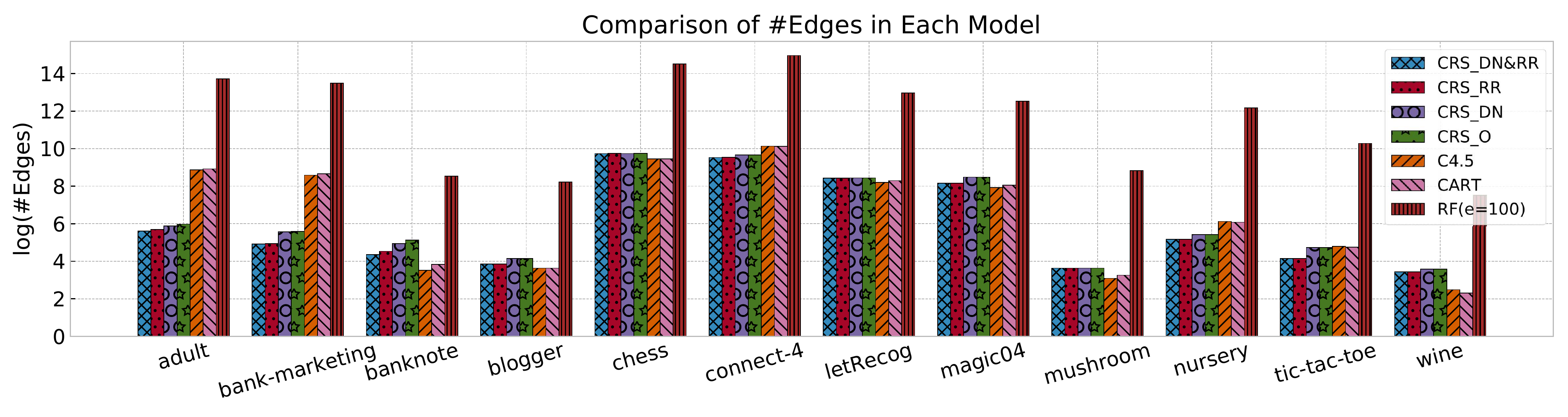}
    \caption{The number of edges in models trained on 12 UCI data sets. The logarithm is adopted for a better viewing experience.}
    \label{fig:model_complexity}
\end{figure*}
\subsubsection{Parameter Settings.}
We set the number of logical layers in CRS, i.e., $2L$, to 4. The number of nodes in each middle layer ranges from 32 to 256 depending on the number of binary features of the data set. The batch size is set to 128, and is trained for 400 epochs. We initialize the learning rate to $5\times10^{-3}$ and decay it by a factor of 0.75 every 100 epochs. The weight decay is set to $10^{-8}$. When using the RB method, we change the selected subset of weights after every epoch and tune the rate of binarization $\mathcal{P}$ using validation sets.

\subsection{Experimental Results}
\subsubsection{Classification Performance}
We first compare the classification F1 score (Macro) of CRS, MLLP trained without RB method (MLLP($\mathcal{P}$=0)), CRS extracted from MLLP trained without RB method (CRS($\mathcal{P}$=0)), C4.5 \cite{Quinlan:1993:CPM:583200}, CART \cite{breiman2017classification}, Scalable Bayesian Rule Lists (SBRL) \cite{yang2017scalable}, Logistic Regression (LR) \cite{kleinbaum2002logistic}, Piecewise Linear Neural Network (PLNN) \cite{chu2018exact}, Support Vector Machines (SVM) \cite{scholkopf2001learning} with linear or RBF kernel, Gradient Boosting Decision Tree (GBDT) \cite{ke2017lightgbm}, and Random Forest \cite{breiman2001random} with 10 estimators (RF(e=10)) and 100 estimators (RF(e=100)).
C4.5, CART and SBRL are all rule-based models, and LR is a linear model. These four models are considered as transparent models and often used as surrogate models. PLNN is a Multilayer Perceptron (MLP) that adopts piecewise linear activation functions, e.g., ReLU \cite{nair2010rectified}. PLNN, SVM, GBDT, and Random Forest are considered as complex models.

The results are presented in Table \ref{tab:accuracy}. We can observe that the average F1 score of CRS on all the data sets is higher than those of other models and CRS outperforms other models on most of the data sets. Only one complex model, i.e., RF(e=100), has a comparable result. However, the Random Forest needs 100 estimators to obtain this result, which is hard to be considered as a transparent model. It should be noted that CRS performs not well on the banknote and letRecog data set. The reason is the recursive minimal entropy partitioning algorithm we applied to discretize continuous feature values brings bias to the data sets. Other models also do not perform well if we train them on these biased data sets. The requirement of data discretization before training is the point we need to improve in future work.

To verify the effect of the RB method, we compare CRS with MLLP trained without the RB method, i.e., MLLP($\mathcal{P}$=0), and the CRS extracted from it, i.e., CRS($\mathcal{P}$=0). We can see that CRS($\mathcal{P}$=0) performs poorly and acts quite differently from its corresponding MLLP, which demonstrates the necessity of the RB method. Moreover, we can see that CRS with RB method outperforms the MLLP trained without the RB method, the reason is the RB method works as the Dropout regularization which significantly alleviates overfitting.

\subsubsection{Impact of Binarization Rate and CRS Structure}
To show the impact of varied random binarization rates $\mathcal{P}$ on CRS with different structures, we train three 3-layer CRS and three 5-layer CRS on the UCI Connect-4 data set using varied binarization rates for illustration. The results are shown in Figure \ref{fig:rb_rate}, and legend labels show the number of nodes in each middle layer, e.g., 64\_64\_64 represents three middle layers and each middle layer has 64 nodes. We can see that without RB method ($\mathcal{P}$=0), deep CRS performs poorly compared to the shallow CRS, and neither shallow CRS nor deep CRS performs well. However, if we set $\mathcal{P}$ to an appropriate value, e.g., ranging from 0.7 to 0.9, the deep CRS can outperform shallow CRS and get a high F1 score. The RB method has no significant influence on the shallow CRS but is very important for the deep CRS. The RB method enables us to train a deeper CRS for higher classification performance. If we set $\mathcal{P}$ too small, the RB method cannot ensure MLLP to keep consistency with CRS. If we set $\mathcal{P}$ too close to 1.0, CRS can hardly be trained well for most of the weights are fixed. Moreover, for CRS with the same depth, the wider one performs better.

\subsubsection{Model Complexity}
Considering that model complexity affects the model transparency, we compare the model complexity of CRS with the decision tree, i.e., C4.5 and CART, and Random Forest, RF(e=100). We also compare the model complexity of CRS after different simplification methods. The metric of model complexity is introduced in the Evaluation Protocols section. The results are shown in Figure \ref{fig:model_complexity}. CRS\_O represents the original extracted CRS without any simplification, CRS\_DN represents CRS using the dead nodes detection method, CRS\_RR represents CRS using the redundant rules elimination method, CRS\_DN\&RR represents CRS using both simplification methods. We can observe that the simplification methods we proposed can reduce the model complexity of CRS on most of the data sets. 
By comparing CRS\_DN\&RR with C4.5 and CART, we can see that CRS trained on adult and bank-marketing data sets have a lower model complexity than C4.5 and CART. On other data sets, there is no significant difference, which verifies the model complexity of CRS is close to decision tree while CRS has a better classification performance. Moreover, Random Forest needs 100 estimators to obtain high classification performance, which leads to extremely high model complexity.
\begin{figure}
    \centering
    \includegraphics[width=0.45\textwidth]{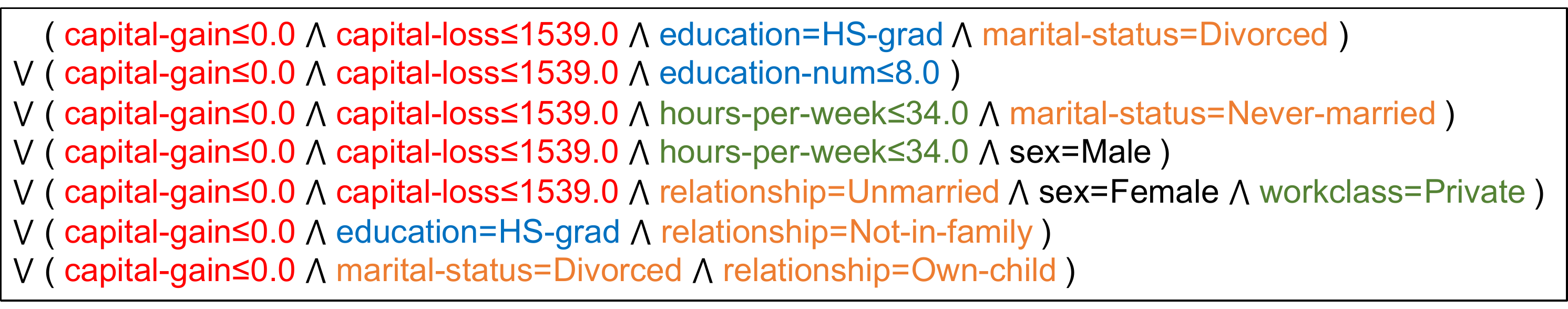}
    \caption{One rule set in CRS trained on the Adult data set.}
    \label{fig:adult_case}
\end{figure}
\subsubsection{Case Study on Inner Structure of CRS}
To show the inner structure of CRS, we select one rule set from the CRS trained on the Adult data set as an example. The rule set shown in Figure \ref{fig:adult_case} corresponds to one CRS node used for predicting whether the income of someone is below \$50K/yr based on census data. Different types of features are marked in different colors, e.g., red for features related to capital. We can clearly see that lacking capital behavior, not being married, short work hours per week and low educational background may lead to low income. Each node in CRS corresponds to one rule or rule set like this example, and we can analyze these nodes to interpret the behavior of the whole model structure.

\section{Conclusion and Future Work}
We propose a new hierarchical rule-based model called Concept Rule Sets and its continuous version, Multilayer Logical Perceptron. We can use gradient descent to learn the discrete CRS efficiently via the continuous MLLP and Random Binarization method we proposed. Our experimental results show that CRS enjoys both high classification performance and low model complexity. 
As future work, we plan to eliminate the need for data discretization before training and explore the application of our method for unstructured data.

\section{Acknowledgments}
 The authors would like to thank the anonymous area chair and (senior) PCs for their valuable comments and suggestions.
 This work is supported by National Natural Science Foundation of China under Grant No. 61532010, 61521002, and 61702190, and Beijing Academy of Artificial Intelligence (BAAI).
 
\bibliographystyle{aaai}
\bibliography{reference} 

\end{document}